# Performance Evaluation of Raster Based Shape Vectors in Object Recognition

Akbar khan[#1], Pratap Reddy L[*2]

[#] *Department of Electronics & Communication Engineering, Nimra Institute of Science & Technology, Vijayawada, India*

[*]*Professor, ECE Department, JNTUH, Hyderabad, India*

*Abstract*— Object recognition is still an impediment in the field of computer vision and multimedia retrieval. Defining an object model is a critical task. Shape information of an object play a critical role in the process of object recognition. Extraction of boundary information of an object from the multimedia data and classifying this information with associated objects is the primary step towards object recognition. Rasters play an important role while computing object boundary. The trade-off lies with the dimensionality of the object versus computational cost while achieving maximum efficiency. In this treatise an attempt is made to evaluate the performance of circular and spiral raster models in terms of average retrieval efficiency and computational cost.

## I. INTRODUCTION

Shape description is the focal step in object recognition playing a paramount role in myriad applications of Machine Vision. Shape description is needed for recognition of 2D flat objects despite deformations in object assuring the detection of objects. The shape is a powerful visual clue for identification and recognition of object. Robust shape features under the influence of deformation can have significant impact on retrieval accuracy [1,8].Many different shape descriptors are proposed by Loncaric.S [2].

The classification of objects is aimed at shape boundary information instead of shape interior information and contour based methods utilize shape boundary information effectively. Fourier descriptor, wavelet descriptor, curvature scale space and shape signatures are some of the contour based shape descriptors. These methods exploit boundary information without capturing interior information which limits to few applications. Region based shape descriptors exploit interior information of a region. This type of object information is also represented by area, treated as global measurement. The area of an object can be determined by counting all pixels with in the object. Raster based shape description provides information about pixels through unique vector for any shape, independent of its position, orientation, or scale. The shape interior information with respect to boundary is represented in the form of vectors. The dimensions of vectors depend on quantization methods. Many researchers attempted to combine area parameter with other parameters to describe shape information. Lu and Sajjanhar [3] proposed grid method. It was applied to contour-based shape, and this convention is followed by Chakrabarti et al.[4]. In grid based shape representation, shape is projected onto a grid of fixed size. Based on the area covered by each cell of grid, a value either 1or 0 is assigned. In this process, a binary sequence is created by scanning the grid from left–to right and top–to bottom which cover the area of object in each cell. This binary sequence is used as shape descriptor while indexing the shape information. Several normalization processes have to be carried out to achieve scale, rotation, and translation invariance. Similarly Goshtasby [5] proposed shape matrix, in which polar raster is used to partition the object. The polar raster consists of concentric circles and radial lines, positioned around centre of the mass. The binary value of the shape is sampled at the intersection of the circles and radial lines. This scheme is invariant to translation, rotation, and scaling. Taza and Suen [6] described shapes by means of shape matrices and a comparison of matrices was performed to classify unknown shapes into one of the known classes. To compensate the difference, weighing parameters are incorporated.

But a shape matrix is a sparse sampling of shape. It is easily affected by noise. Besides, shape matching using a shape matrix is very expensive. Later Perui et al [7] proposed a shape description scheme based on the relative areas of the shape contained in concentric rings located in the shape centre of the mass. The shape matrix technique captures local shape information well, but it suffer from noise and high dimension matching. The area-ratio technique is robust to noise, but it does not capture local shape information by computing area in each ring.

Zhang and Lim [9] proposed a polar raster to extract local information by sampling signature technique to compute a signature function of the sampled points using circular raster. Few more raster models using spiral and square shapes are proposed by Khan& Reddy [10]. In all these models, instead of computing whole region, raster models are used for sampling at regular intervals. Shape pixels are counted at these sampling points. Using these raster models radial vectors and angular vectors are derived. Shape comparison is done using these shape vectors. The 0 s and 1s in the vectors represent points that belong to the outermost and inner most regions of the shape respectively. This technique provides the information about pixels through unique vector for any shape independent of its position, orientation or scale. The shape interior information with respect to boundary is represented in the form of vectors in all the presented methods. The dimension of these vectors depends on the quantization methods that are used. In this paper the performance of various raster models are compared for object recognition in terms of retrieval efficiency, and computational cost. The effect of shape vector length and sampling rate on retrieval efficiency and computational complexity are studied for various raster models.

## II. Shape Information Using Rasters

Object boundary is important information while extracting the shape. Identification of exterior and interior parts of boundary is the complex step involved in object identification. Region based representation explores spatiotemporal characteristics of the object. Grid based region extraction methods are computationally intensive while covering the interior region of the object. Exterior region of boundary also play a vital role while predicting the boundary. In this context, two major approaches can be attempted using rasters. The first approach computes the area covered by the object boundary using radial direction. A raster will be super-imposed on the object. The raster is dissected through either uniformly or non-uniformly spaced samples. Each sample will be identified with its presence related to interior portion of the object boundary or exterior region. All samples that are present in the respective regions will provide shape information of that object. To acquire more precise information, the number of raster cycles and samples can be increased which results in a radial vector of raster. Similarly second approach can be adopted for the same raster using angular dissection. The dissected region of the raster with predefined angle will provide the boundary information in terms of samples that reside within the interior region or placed outside the boundary giving rise to exterior information. The dissection in angular direction can be carried out with uniform or non- uniform approach. The quantified samples of each angularly segmented region will result in angular vector of the shape. In both approaches, the entire area will be computed by predicting the presence of sampled points within or outside the boundary. However the trade-off lies in both approaches in terms of number of cycles (segments) that are necessary. In this paper, rigorous efforts are made to evaluate trade-off between cycles per raster versus samples in each cycle, where each cycle represent the segmented region of the object.

### A. Circular raster

Circular raster, proposed by khan [10] consists of concentric circles along with radial lines. The circles and radial lines are sampled at regular intervals. For computation of radial and angular segment vectors, the circular raster is laid over the shape such that its centroid coincides with centre of the raster. Two types of shape vectors are proposed. First, radial vector is defined as the number of samples present on each circle in the raster, later normalized with the size of raster. In the other approach, angular vector is defined as number of samples present on each radial line at regular fixed angle on circular raster. The circular raster and corresponding shape vectors of the associated object are presented in Fig. 1

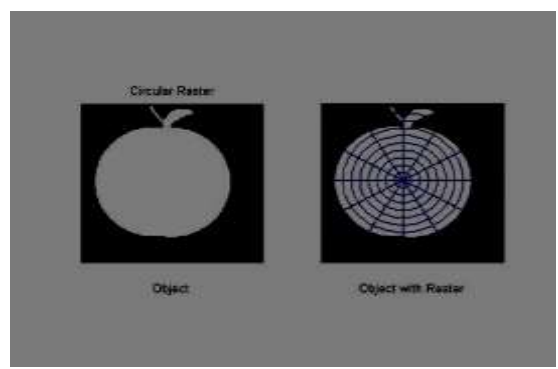

(a)

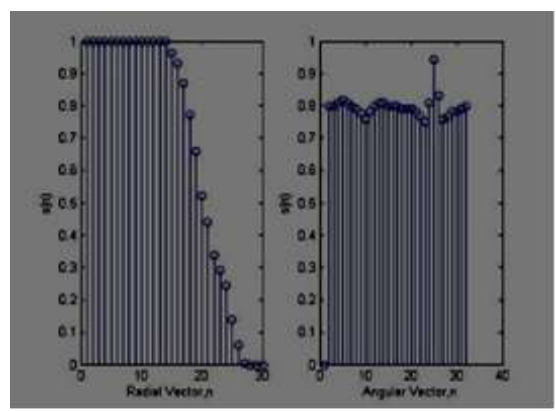

(b)

Fig .1 (a) Object and Circular raster overlaid on object (b) Radial shape Vector and Angular shape Vector.

*B. Archimedean Spiral Raster*

Archimedean spiral along with radials at regular intervals forms spiral raster. In Archimedean spiral the distance between successive turnings is constant and radius varies with angle. Two types of shape radial vectors are proposed, namely Full cycle and Fixed angle radial shape vector. In full cycle Radial Shape Vector, a full cycle is composed by spiral at centre, considered as full cycle, i.e. 360 degrees. Number of cycles is considered on spiral to cover query object. Radial shape vector is computed as total number of image sample pixels for one full cycle. The radial vector is normalized with respect to total pixels on the cycle. The length of radial vector depends on separation between cycles. Similarly in fixed angle, radial vector spiral raster is divided into segments at a fixed angle. For computing shape radial vector, shape pixels are counted on each spiral segment between two radial arms. The shape vector with full cycle and fixed cycle using Archimedean spiral are presented in Fig. 2 & 3 respectively.

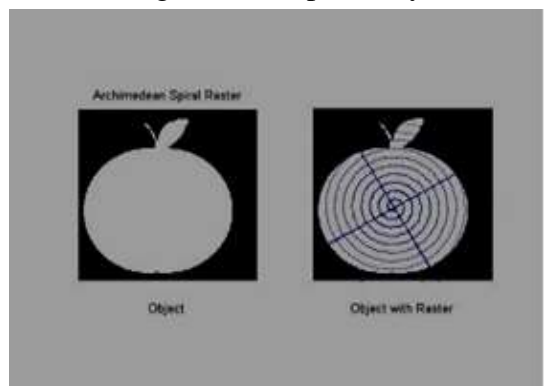

(a)

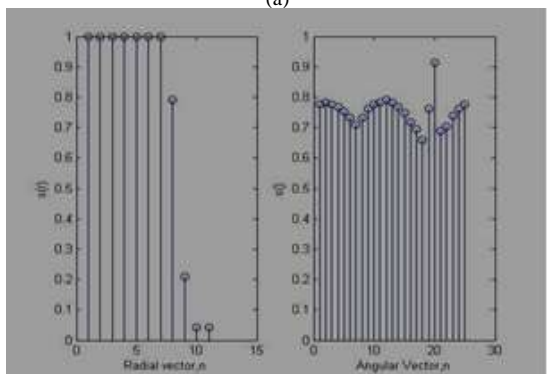

(b)

Fig.2 (a) Object and Archimedean spiral raster (b) Radial shape vector and Angular shape vector using full cycle.

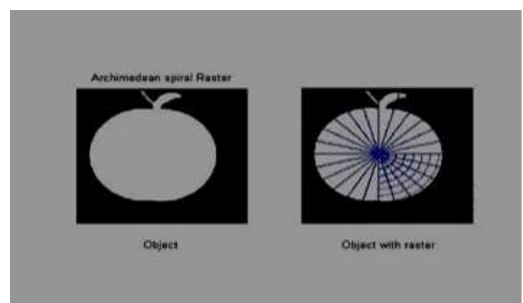

(a)

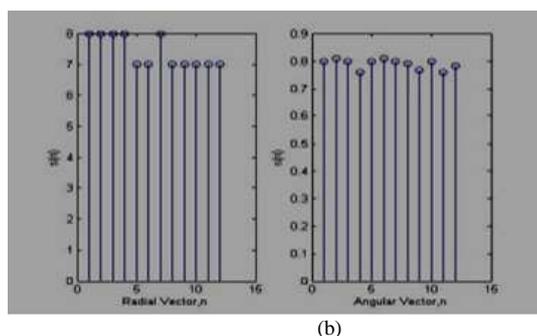

Fig. 3 a) Object and Archimedean spiral raster b) radial shape vector and Angular shape vector Using fixed cycle.

### III. OBJECT IDENTIFICATION

The proposed approach of predicting object shape with the help of raster has a trade-off in terms of practical number of cycles in radial direction as well as angular direction. In ideal scenario the maximum number of cycles is equal to the length of spiral from centroid to the peak of the boundary in radial direction. In angular direction the maximum number of segments will be infinity. It is necessary to estimate the trade-off between the maximum and minimum cycles. In the present approach we attempted to evaluate this trade-off with respect to the associated computational time. We adopted supervised training approach for this purpose. In training phase 23 objects are considered for evaluation with each object possessing 20 images with varied orientation, size and translation. The proposed raster is superimposed on each object. The radial and angular vectors are computed as discussed in section 2 while providing the shape information. In testing phase the classification is carried out with the help of Euclidean distance between the test vectors and trained database.

### IV. PERFORMANCE EVALUATION

In this work, the first evaluation criterion is associated with average retrieval efficiency which is defined as the ratio of number relevant objects recognized in each category to the total number of test objects. The second evaluation parameter is taken as computation time. Even with available high speed computing facilities, accurate and fastest algorithms are always needed forever. This is particularly true in the field of object recognition where a multitude of applications enforce real time computation. It is hard to compare different recognition methods using this criterion because the computational time strongly depends on the individual implementation of recognition methods. Hence, the average computational time for an object is considered as ratio of time taken for retrieval of total test objects with number of test objects. Further the computation time and efficiency depends on shape vector length or resolution of the vector respectively. Resolution also influences these parameters. The trade-off between sampling rate and size of raster models are considered in terms of efficiency and complexity cost. The other measurement considered here is associated with robustness of these models. In this work robustness is measured based on occlusion and also introduced occlusion in the test set as a sample set. The robustness for various raster models at critical sampling and raster sizes are computed and compared and analysed.

A test dataset of 460 images from MPEG-7 CE shape-1 part B consists of 23 categories with 20 samples is chosen for performance evaluation. Radial shape vector for all test objects is computed using above mentioned rasters and stored in different databases. For recognition of query objects the radial vectors of query objects are computed using circular raster and compared with radial shape vectors of same raster. To compare shape vectors Euclidean distances are computed and objects related to first three minima distance are considered as closely matching objects. Same procedure is repeated for recognition of remaining rasters. The length of radial shape vector depends on size of raster, which in turn depends on number of circles or cycles in raster. Also the resolution of radial shape vector depends on sampling rate. So these two factors affect the retrieval efficiency, computation time and robustness. Circular raster object recognition is performed with various raster sizes such as varying the distance between cycles in circular raster and number of samples on cycles also varied. All these raster are evaluated with respect to average retrieval efficiency and computation time for all test objects and computed results are discussed in detail.

## V. RESULTS & DISCUSSIONS

With circular raster the average retrieval efficiencies with different distances between cycles and at various sampling rates per cycle are presented in Table 1 with associated graph in Fig 5. It is observed from the results, with circular raster, average retrieval efficiencies increases with sampling rates and also increasing with reduction in distance between cycles. At the same time the retrieval time for circular raster at various sampling rates and with various separations between cycles are provided in Table 2 and Fig 6. It is observed that retrieval time increases with samples as well as reduction in distance between cycles. In case of circular raster using radial shape vector the retrieval efficiency is nearly equal for both cases with 24 pixels separation between cycles with 24 samples per cycle and 8 pixels separation with 6 samples per cycle. But the computation time is relatively higher with 8 pixels separation and 6 samples. Object Recognition is performed for all test objects using full cycle shape vectors for various distances between cycles and sampling rate per cycle with Archimedean spiral raster. The results are presented in Table 3, and in Fig 7. The retrieval time is listed in Table 4 and Fig 8. For Archimedean spiral raster with full cycle the average retrieval efficiency is approximately same with 32 pixels separation with 24 samples and 8 pixels separation with 6 samples. Whereas the computation time is relatively small in first case when compared with second one. Further with fixed cycle shape vectors object recognition is performed and the retrieval efficiency and computational time are presented Tables5 &6 and corresponding graphs are shown in Figs. 9 & 10. With fixed cycle shape vectors in Archimedean spiral, retrieval efficiency is comparatively similar in three cases i.e., 8 pixels at 6 samples, 16 pixels at 8 sample and 32 pixels at 24 samples. When compared with computation time significant difference is observed with 16 samples at 8 samples. The computing procedure of angular shape vector is same for all raster models. Computation using Angular vectors is performed on test objects at various sizes and sampling rates. The results of retrieval efficiency and computation time are shown in Tables 7& 8 and Figs 11 & 12. In case of angular shape vectors the relative efficiency and time are observed with 16 pixels separation and with 8 samples. For measuring robustness, retrieval rates for these shape vectors are computed using permitted occluded objects test set consist 46 objects 2 from each category and presented in Fig.13. The retrieval efficiency of these occluded objects is performed with circular raster, Archimedean spiral raster with full cycle and fixed cycle for above discussed sampling and separations between cycles only. These results are tabulated in Table 9. It is observed that robustness is more for full cycle radial shape vectors using Archimedean spiral raster and robustness is same for radial shape vectors and angular shape vector in circular raster.

TABLE I. AVERAGE Retrieval efficiency for CIRCULAR RASTER RADIAL VECTOR

| S.No | Separation between cycles | 4 samples per cycle | 6 samples per cycle | 8 samples per cycle | 12 samples per cycle | 24 samples per cycle | Name of the curve |
|---|---|---|---|---|---|---|---|
| 1 | 32 | 36 | 61 | 66 | 79 | 93 | Series1 |
| 2 | 24 | 57 | 72 | 82 | 84 | 96 | Series 2 |
| 3 | 16 | 65 | 82 | 89 | 92 | 99 | Series3 |
| 4 | 8 | 84 | 95 | 95 | 96 | 100 | Series4 |

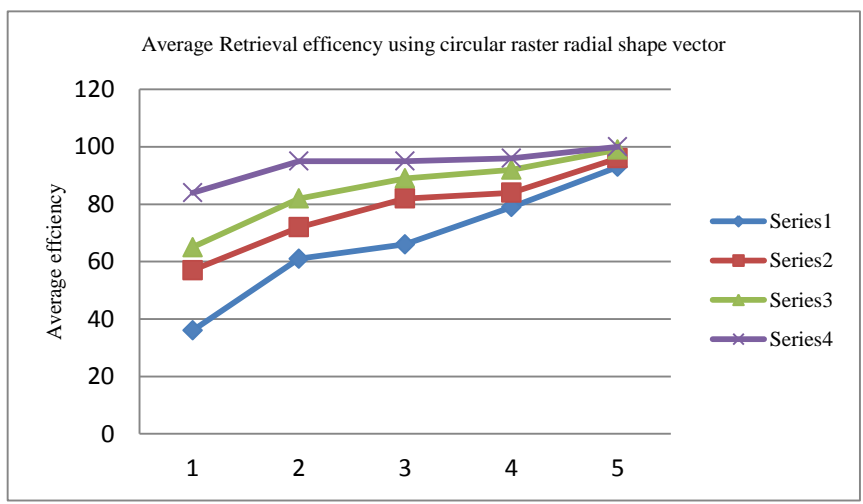

Fig 5. Average Retrieval efficiency using circular raster radial shape vector

TABLE II. TOTAL RETRIEVAL TIME WITH CIRCULAR RASTER RADIAL SHAPE VECTOR

| S.No | Separation between cycles | 4 samples per cycle | 6 samples per cycle | 8 samples per cycle | 12 samples per cycle | 24 samples per cycle | Name of the curve |
|---|---|---|---|---|---|---|---|
| 1 | 32 | 108.12 | 104.87 | 105.03 | 104.63 | 104.83 | Series1 |
| 2 | 24 | 110.39 | 106.58 | 106.22 | 113.83 | 108.68 | Series 2 |
| 3 | 16 | 107.03 | 114.29 | 110.05 | 118.97 | 118.97 | Series3 |
| 4 | 8 | 120.88 | 121.8 | 122.59 | 125.64 | 120.41 | Series4 |

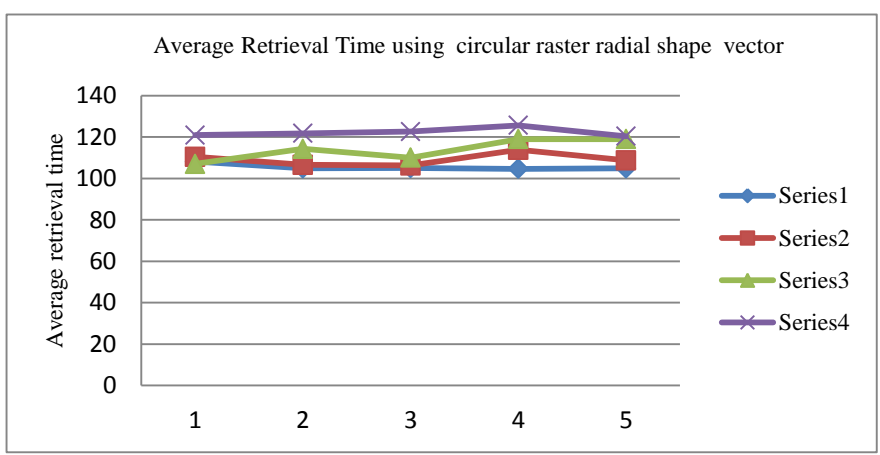

Figure. 6. Total retrieval time with circular radial vector

TABLE III. AVERAGE RETRIEVAL EFFICIENCY WITH ARCHIMEDEAN SPIRAL RASTER RADIAL VECTOR WITH FULL CYCLE

| S.No | Separation between cycles | 4 samples per cycle | 6 samples per cycle | 8 samples per cycle | 12 samples per cycle | 24 samples per cycle | Name of the curve |
|---|---|---|---|---|---|---|---|
| 1 | 32 | 30 | 54 | 70 | 82 | 96 | Series1 |
| 2 | 24 | 54 | 70 | 80 | 87 | 95 | Series 2 |
| 3 | 16 | 69 | 83 | 88 | 93 | 98 | Series3 |
| 4 | 8 | 89 | 95 | 97 | 98 | 100 | Series4 |

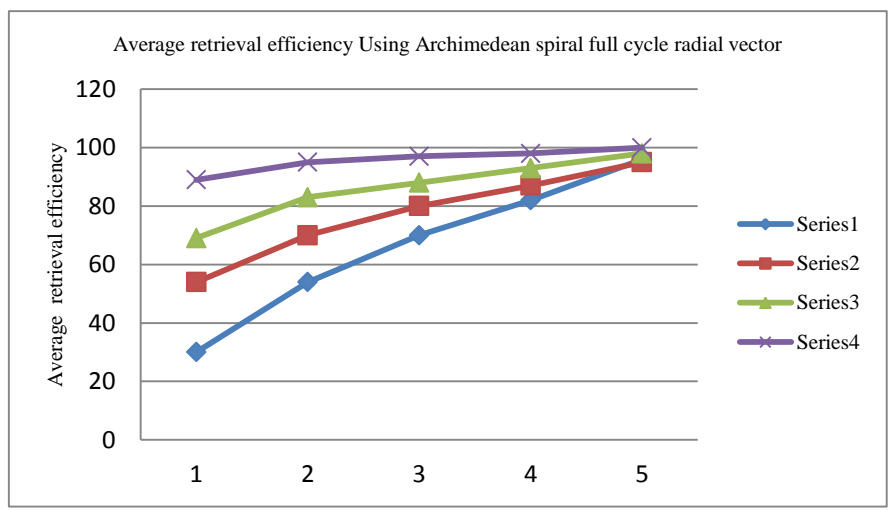

Fig. 7. Average Retrieval efficiency with Archimedean spiral Raster radial vector with full cycle

TABLE IV. RETRIEVAL TIME WITH ARCHIMEDEAN SPIRAL RASTER RADIAL SHAPE VECTOR WITH FULL CYCLE

| S .No | Separation between cycles | 4 samples per cycle | 6 samples per cycle | 8 samples per cycle | 12 samples per cycle | 24 samples per cycle | Name of the curve |
|---|---|---|---|---|---|---|---|
| 1 | 32 | 104.295 | 104.49 | 104.68 | 105.5 | 104.66 | Series1 |
| 2 | 24 | 106.705 | 106.25 | 106.3 | 106.24 | 106.58 | Series 2 |
| 3 | 16 | 109.203 | 110.06 | 109.13 | 105.9 | 106.99 | Series3 |
| 4 | 8 | 123.003 | 117.06 | 116.66 | 117.75 | 118.18 | Series4 |

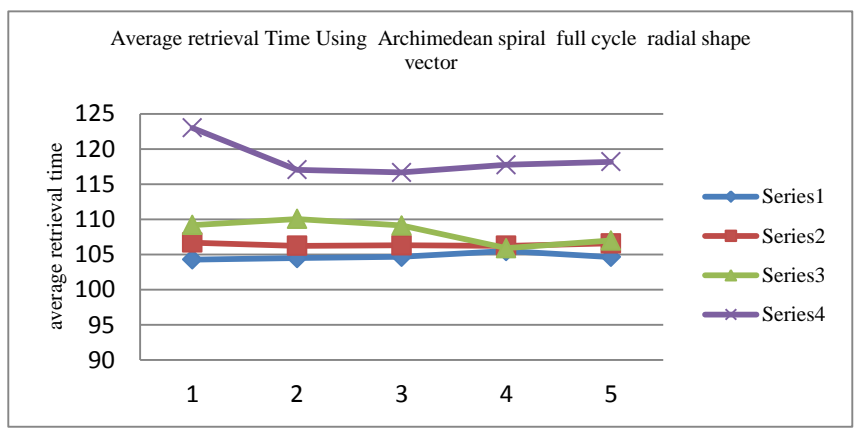

Figure. 8.   Retrieval time with Archimedean spiral raster   radial shape vector with full cycle.

TABLE V. AVERAGE RETRIEVAL EFFICIENCY WITH ARCHIMEDEAN SPIRAL RASTER RADIAL VECTOR WITH FIXED CYCLE

| S.No | Separation between cycles | 4 samples per cycle | 6 samples per cycle | 8 samples per cycle | 12 samples per cycle | 24samples per cycle | Name of the curve |
|---|---|---|---|---|---|---|---|
| 1 | 32 | 81 | 87 | 91 | 92 | 100 | Series1 |
| 2 | 24 | 85 | 90 | 95 | 94 | 99 | Series 2 |
| 3 | 16 | 92 | 96 | 98 | 97 | 99 | Series3 |
| 4 | 8 | 96 | 98 | 98 | 98 | 100 | Series4 |

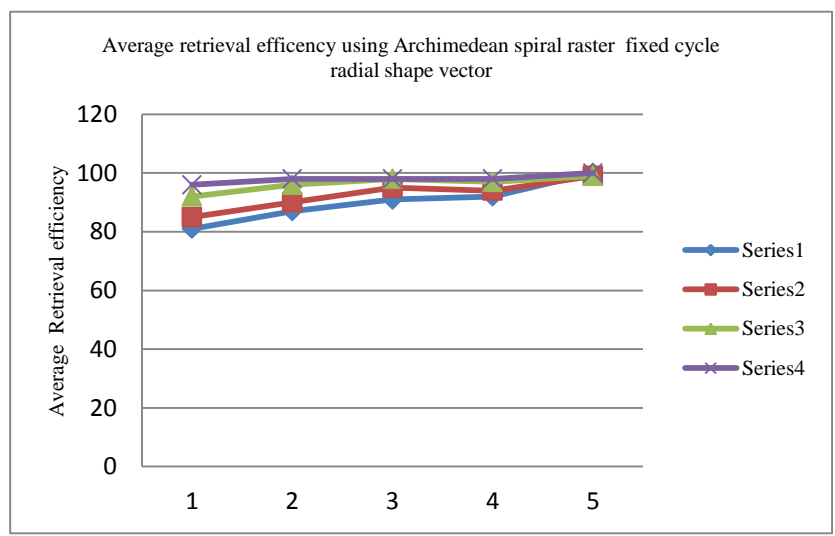

Figure. 9.  Average Retrieval efficiency using Archimedean   spiral fixed cycle raster radial vector

TABLE VI. RETRIEVAL TIME USING ARCHIMEDEAN SPIRAL RASTER RADIAL SHAPE VECTOR WITH FIXED CYCLE

| S.no | separation between cycles | 4 samples per cycle | 6 samples per cycle | 8 samples per cycle | 12 samples per cycle | 24 samples per cycle | Name of the curve |
|---|---|---|---|---|---|---|---|
| 1 | 32 | 103.68 | 103.95 | 104.03 | 103.62 | 111.38 | Series1 |
| 2 | 24 | 115.56 | 108.84 | 112.01 | 107.78 | 103.77 | Series 2 |
| 3 | 16 | 105.66 | 105.02 | 105.49 | 112.48 | 109.87 | Series3 |
| 4 | 8 | 105.54 | 105.83 | 114.25 | 104.21 | 105.78 | Series4 |

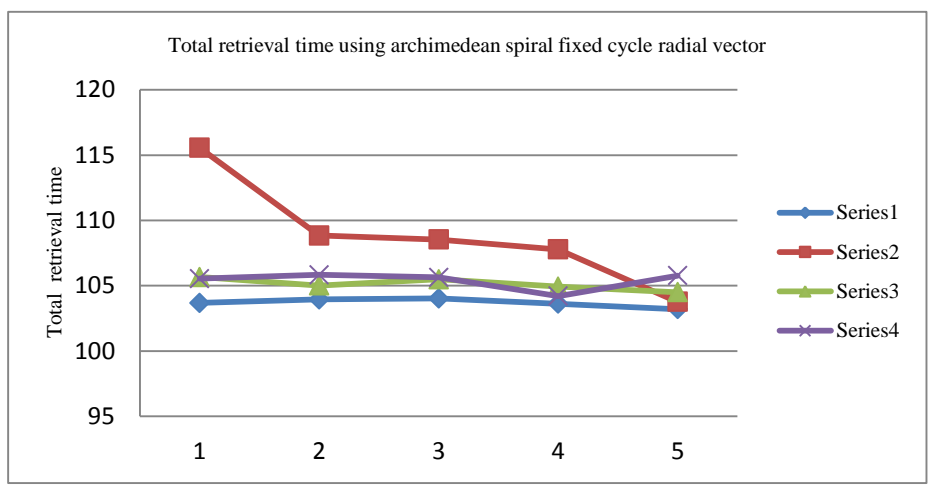

Figure 10. Retrieval time using Archimedean spiral Raster radial shape vector with fixed cycle

TABLE VII. AVERAGE RETRIEVAL EFFICIENCY USING ANGULAR SHAPE VECTOR

| S.No | Separation between cycles | 4 samples per cycle | 6 samples per cycle | 8 samples per cycle | 12 samples per cycle | 24 samples per cycle | Name of the curve |
|---|---|---|---|---|---|---|---|
| 1 | 32 | 93 | 95 | 96 | 98 | 100 | Series1 |
| 2 | 24 | 95 | 95 | 97 | 97 | 99 | Series 2 |
| 3 | 16 | 95 | 96 | 97 | 98 | 100 | Series3 |
| 4 | 8 | 93 | 98 | 98 | 99 | 100 | Series4 |

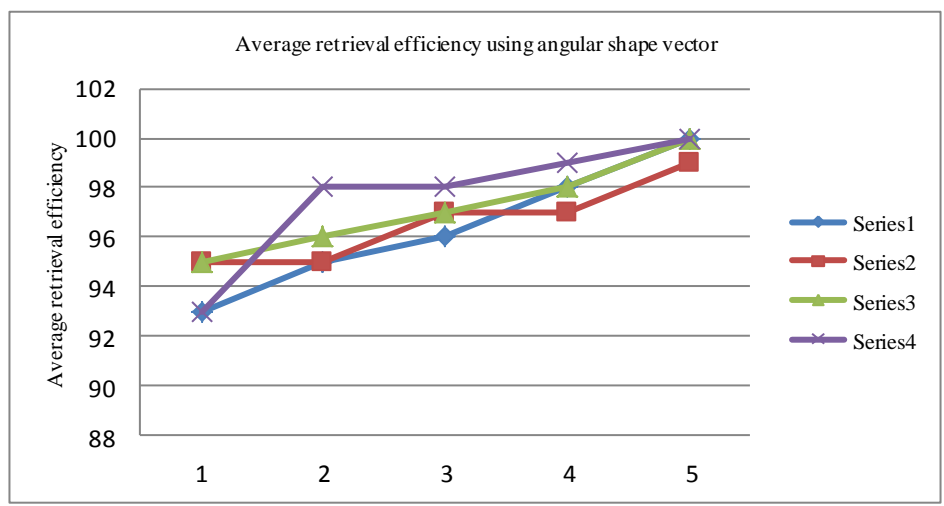

Figure. 11. Average Retrieval efficiency with Angular vector

TABLE VIII. RETRIEVAL TIME USING ANGULAR SHAPE VECTOR

| S.No | Separation between cycles | 4 samples per cycle | 6 samples per cycle | 8 samples per cycle | 12 samples per cycle | 24 samples per cycle | Name of the curve |
|---|---|---|---|---|---|---|---|
| 1 | 32 | 126.09 | 127.89 | 132.33 | 132.61 | 145.55 | Series1 |
| 2 | 24 | 106.13 | 108.07 | 124.54 | 131.49 | 146.4 | Series 2 |
| 3 | 16 | 105.69 | 113.22 | 112.22 | 113.98 | 125.42 | Series3 |
| 4 | 8 | 107.48 | 113.17 | 109.72 | 113.31 | 125.47 | Series4 |

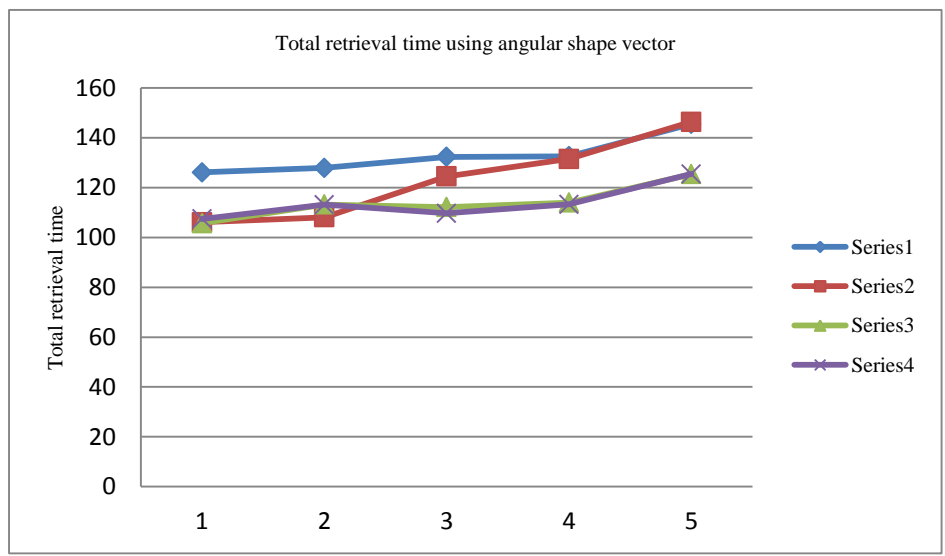

Figure. 12. Total retrieval time using Angular shape vector

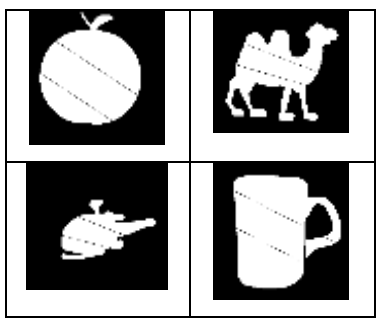

Fig 13. Occluded test objects

TABLE IX. RETRIEVAL EFFICIENCY OF OCCLUDE OBJECTS

| S. No | Name of raster | Average Retrieval efficiency |
|---|---|---|
| 1 | Circular raster radial shape vector 24 pixels at 24 samples | 86.9 % |
| 2 | Archimedean spiral raster radial vector with full cycle 32 pixels at 24 samples | 95.6 % |
| 3 | Archimedean spiral raster with fixed cycle 24 pixels at 12 samples | 48.7% |
| 4 | Angular vector 16 pixels at 8 samples | 86.9% |

VI. CONCLUSIONS

In this exiting treatise, the researchers succeeded to assess the performance of circular, spiral based shape raster using radial and angular function. Object model is presented using boundary with respect to core and peripheral information of a specific segment correlating with the maximum distance of the object from the centroid. In this process the trade-off between radial and angular segments of the raster is computed and evaluation of performance is carried out to achieve 100% retrieval efficiency. We observed 8 pixels per cycle as the optimum distance with 24 samples per cycle for circular raster. A comparable computational cost is observed. Similar observations are made with Archimedean spiral raster also. However close to the maximum retrieval efficiency is observed with Archimedean spiral with 12 samples per cycle. In these conditions, occluded objects recognition using angular vector of spiral raster is found to be more efficient. The performance evaluation trade-off issues with the help of combined features of radial and angular vectors are in progress.